\documentclass{article}

\usepackage{microtype}
\usepackage{graphicx}
\usepackage{subfigure}
\usepackage{booktabs}
\usepackage{xcolor} 

\usepackage{hyperref}

\usepackage{makecell}
\usepackage{amsfonts}
\usepackage{amsmath}
\usepackage{bbm}
\usepackage{subfigure}
\usepackage{bigstrut}
\usepackage{colortbl}

\usepackage[accepted]{mlsys2023}

\newcommand{\R}{\mathbb{R}}

\newtheorem{takeaway}{Takeaway}

\definecolor{Gray}{gray}{0.95}
\definecolor{DarkGray}{gray}{0.5}
\definecolor{LightCyan}{rgb}{0.88,1,1}

\begin{document}

\twocolumn[

\mlsystitle{Does compressing activations help model parallel training?}



\mlsyssetsymbol{equal}{*}

\begin{mlsysauthorlist}
\mlsysauthor{Song Bian}{equal,wisc}
\mlsysauthor{Dacheng Li}{equal,cmu}
\mlsysauthor{Hongyi Wang}{cmu}
\mlsysauthor{Eric P. Xing}{cmu,MBZUAI,petuum}
\mlsysauthor{Shivaram Venkataraman}{wisc}
\end{mlsysauthorlist}

\mlsysaffiliation{wisc}{Department of Computer Science, University of Wisconsin-Madison}
\mlsysaffiliation{MBZUAI}{MBZUAI}
\mlsysaffiliation{petuum}{Petuum Inc.}
\mlsysaffiliation{cmu}{Machine Learning Department, Carnegie Mellon University}

\mlsyscorrespondingauthor{Song Bian}{songbian@cs.wisc.edu}

\mlsyskeywords{Machine Learning, MLSys}

\vskip 0.3in

\begin{abstract}
Large-scale Transformer models are known for their exceptional performance in a range of tasks, but training them can be difficult due to the requirement for communication-intensive model parallelism. One way to improve training speed is to compress the message size in communication. Previous approaches have primarily focused on compressing gradients in a data parallelism setting, but compression in a model-parallel setting is an understudied area. We have discovered that model parallelism has fundamentally different characteristics than data parallelism. In this work, we present the first empirical study on the effectiveness of compression methods for model parallelism. We implement and evaluate three common classes of compression algorithms - pruning-based, learning-based, and quantization-based - using a popular Transformer training framework. We evaluate these methods across more than 160 settings and 8 popular datasets, taking into account different hyperparameters, hardware, and both fine-tuning and pre-training stages. We also provide analysis when the model is scaled up. Finally, we provide insights for future development of model parallelism compression algorithms. 

\end{abstract}
]



\printAffiliationsAndNotice{\mlsysEqualContribution} 

\section{Introduction}
\label{sec:intro}

\begin{figure}[!ht]
\centering
\includegraphics[width=0.4\textwidth]{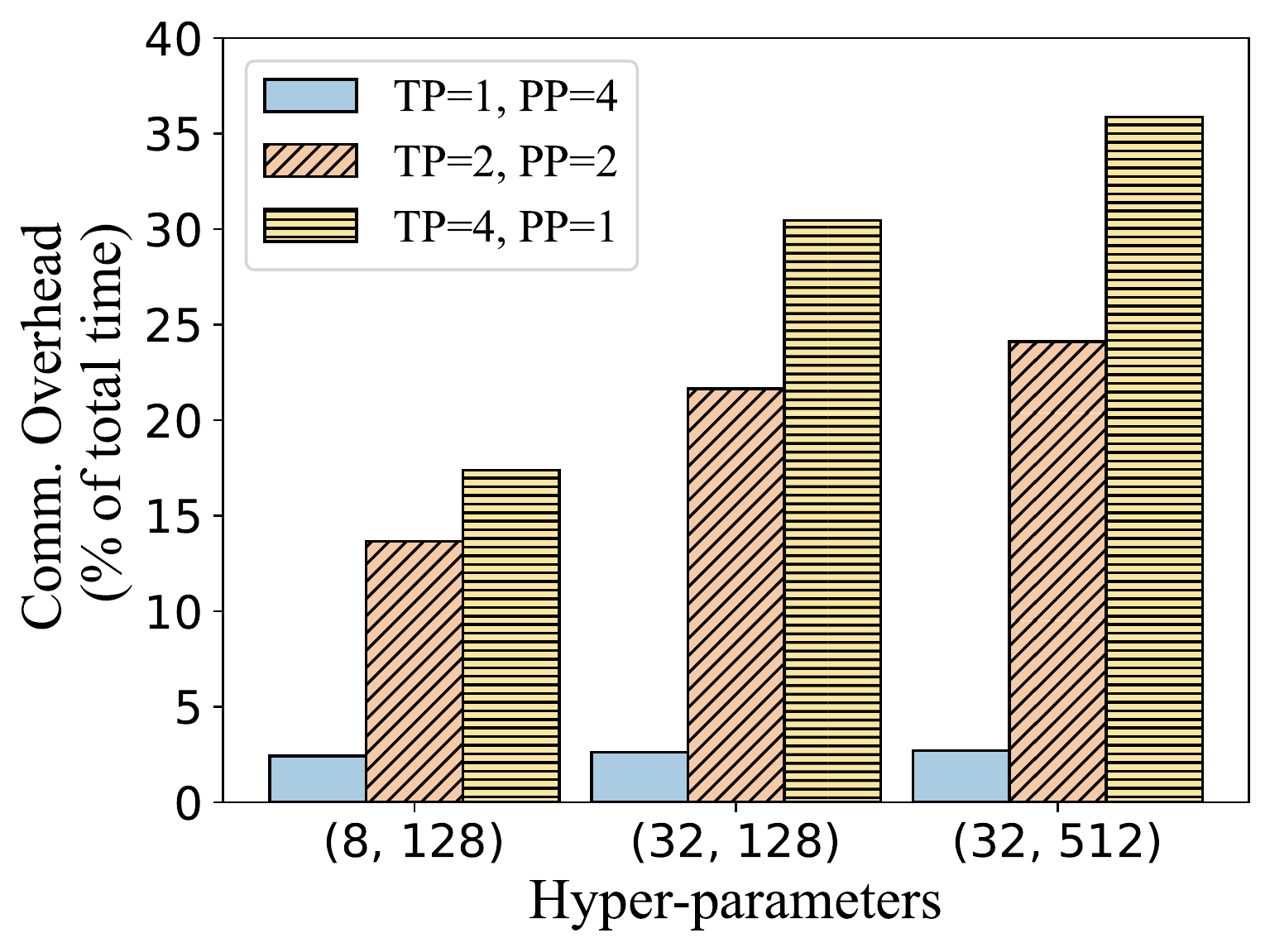}
\vspace{-2mm}
\caption{
Communication overhead of model parallelism with different batch sizes and sequence lengths on $\text{BERT}_{\text{Large}}$ using Pytorch 1.12, NCCL, fp16 and 4 GPUs. The $x$-axis is (batch size, sequence length)}
\label{fig:comm_cost}
\vspace{-4mm}
\end{figure}

Transformer models have become the dominant model for many machine learning tasks~\cite{devlin2018bert,radford2018improving,yang2019xlnet, dosovitskiy2020image,gong2021ast,sharir2021image, gong2021ast}. However, state-of-the-art Transformer models have a large number of parameters, making it difficult for a single GPU to hold the entire model. As a result, training large Transformer models often requires partitioning the model parameters among multiple GPUs, a technique known as {\it model parallelism}~\cite{shoeybi2019megatron, rasley2020deepspeed}. Model parallelism strategies often introduce significant communication overhead, as demonstrated in Figure~\ref{fig:comm_cost}~\cite{li2022amp}. For instance, the most commonly used tensor model parallelism strategy requires two all-reduce operations over a large tensor in {\bf each} Transformer encoder block per iteration. This can greatly increase the overall computational cost of training the model~\cite{shoeybi2019megatron}. 


\begin{figure}[!t]
\centering
\includegraphics[width=0.4\textwidth]{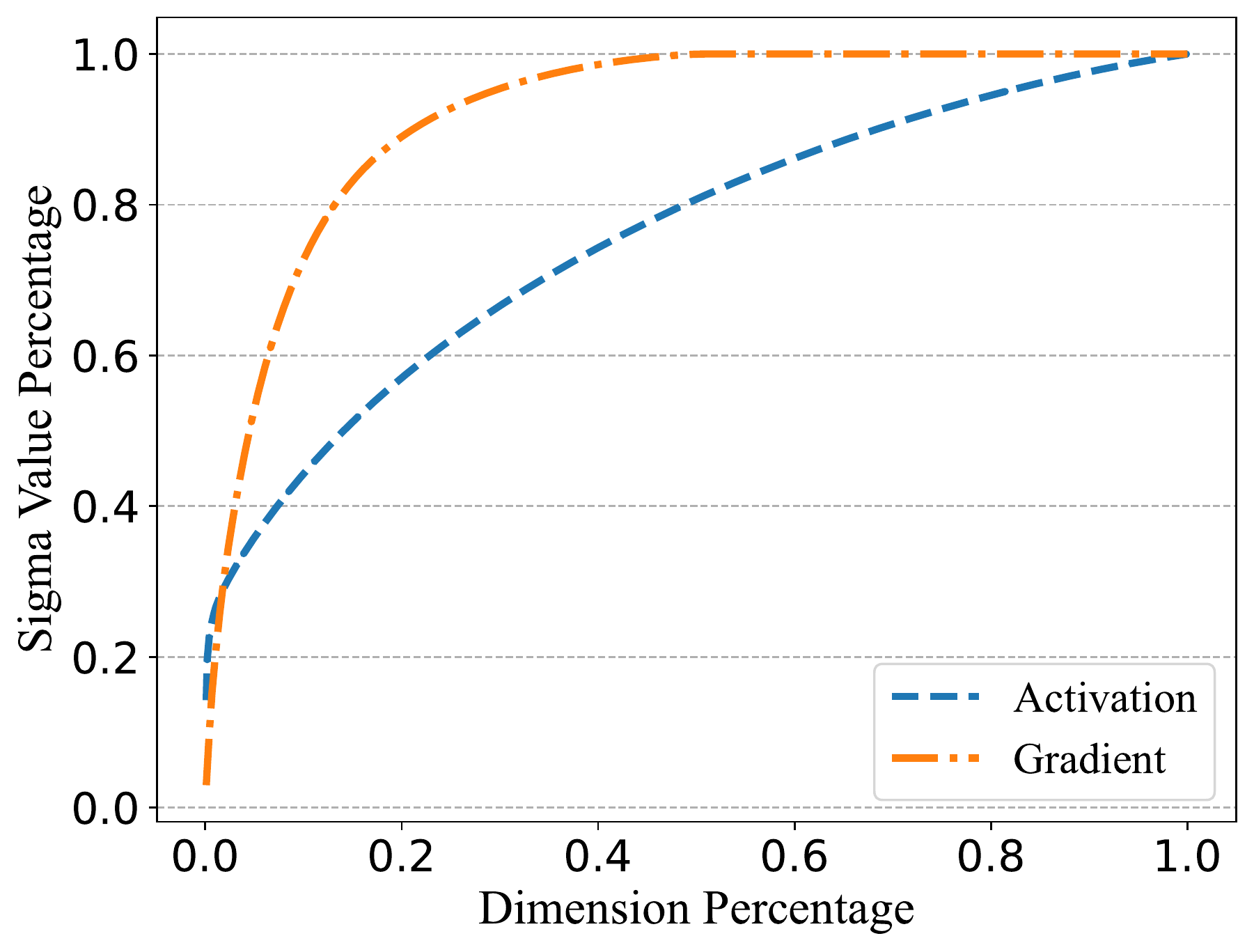}
\vspace{-1mm}
\caption{\textbf{Low-Rank analysis:} Curves are drawn by ordering the singular values of the SVD decomposition. The result shows that the gradient is low-rank but the activation is not. The activation is the output of the 12$^{th}$ transformer layer in $\text{BERT}_{\text{Large}}$ model.}
\vspace{-2mm}
\label{fig:low_rank}
\end{figure}

To address the issue of high communication overhead in model parallelism, one approach is to compress the messages communicated among GPUs, such as activation values. In the data-parallel setting, several prior works have explored compressing gradients to reduce the communication cost of training~\cite{seide20141, bernstein2018signsgd,dettmers20158, lin2017deep, wang2018atomo,vogels2019powersgd}.
However, there has been limited exploration of compression methods specifically designed for model parallelism. Furthermore, it is important to note that compression in model parallelism is fundamentally different from compression in data parallelism for two main reasons. Firstly, as shown in Figure~\ref{fig:low_rank}, gradients tend to be low-rank, while activations do not. Therefore, low-rank gradient compression methods, which have been shown to provide state-of-the-art end-to-end speedup in communication-efficient data-parallel training, may not directly apply to model parallelism~\cite{vogels2019powersgd}. 
Secondly, the performance benefits of gradient compression methods can be significantly affected by system optimizations in data parallelism~\cite{agarwal2022utility}. However, model parallelism has a different set of system optimization techniques than data parallelism, so it is unclear how these optimizations would impact the performance of compression methods in model parallelism. 

In this paper, we present the first systematic study of model parallelism compression for large Transformer models. We evaluate the impact of different compression methods in terms of both {\bf throughput} and {\bf accuracy}. We conduct experiments for both pre-training and fine-tuning tasks.
~\cite{devlin2018bert, gururangan2020don}. In particular, we implement and evaluate popular gradient compression methods, e.g., Top-$K$ and Random-$K$ as well as a learning-based compression method, i.e.,  auto-encoders~\cite{hinton1993autoencoders}, which can not directly be applied to gradient compression but is compatible with activation compression. 
To assist researchers and practitioners training new Transformer-based models~\cite{liu2019roberta, izsak2021train}, we study compression methods using different training hyper-parameters and hardware setups. We also develop a performance model that can be conveniently used 
to understand how compression methods would affect throughput at larger scales. In total, we evaluate compression methods across over 160 different settings with various compression algorithms, training stages, hyper-parameters, and hardware, and over 8 datasets~\cite{wang2018glue}. Our findings include the following takeaways.

\textbf{Our takeaways.} \textbf{1. Learning-based compression methods are most suitable for model-parallelism.} On the fine-tuning stage(\S \ref{sec:throughput_ft}, \S \ref{sec:accuracy_ft}), only auto-encoders (AEs) 
can provide end-to-end speedup (upto \textbf{18\%}) while preserving the model's accuracy (within $\sim$\textbf{3} GLUE score~\cite{wang2018glue}). Top-$K$, Random-$K$, and quantization methods slow down training because their message encoding and decoding overhead is larger than the communication time they reduce. Top-$K$ and Random-$K$ also hurt model's accuracy.

For the pre-training stage (\S \ref{sec:throughput_pt}), only AE 
provides speedup (upto \textbf{16\%}) while preserving the model's accuracy (\textbf{similar} GLUE score). Top-$K$ marginally improves training time, but degrades the accuracy. Quantization slows down the training time, and degrades the accuracy.

\textbf{2. Training hyper-parameters affect the performance benefits of compression methods.} None of the compression methods can improve performance when the batch size and sequence length are small because the cost of message encoding and decoding becomes relatively higher (as discussed in section \S \ref{sec:impact_hyper}). In practice, we have found that the batch size and sequence length need to be at least 32 and 512, respectively, for the compression methods to provide throughput gains. The same is true when fine-tuning is performed on a machine with high-bandwidth NVLink connections between all GPUs (as described in section \S\ref{sec:throughput_ft}). 

\textbf{3. Early model layers are more sensitive to compression.} 
Our observations show that compressing the early layers or too many layers significantly decreases the model's accuracy (as discussed in section \S\ref{sec:impact_times_location}), which is consistent with the findings of previous research~\cite{wang2021pufferfish}. In practice, we have found that compressing the final 12 layers of a 24-layer Transformer model is an effective approach.
\textbf{Contributions.} We make the following contributions:
\begin{itemize}
    \item We conduct the first empirical study on model parallelism compression methods for Transformer models, considering different compression methods, training stages, hyper-parameters, and hardware configurations. 
    \item We implement several popular compression algorithms, including Top-$K$, Random-$K$, quantization, and auto-encoders (AEs), and integrate them into an existing distributed training system.
    \item We extensively evaluate these algorithms across over 160 different settings and eight popular datasets. Based on our experimental results, we provide several takeaways for future model parallelism compression studies. We also analyze the speedup when the model size and cluster size are scaled up.
    
\end{itemize}

\section{Background and Challenges}
\label{sec:background}
In this section, we first introduce data parallelism and model parallelism (\S \ref{sec:dp_mp}). Then we introduce the challenges in model parallelism compression (\S \ref{sec:challenge}).

\subsection{Data Parallelism and Model Parallelism}
\label{sec:dp_mp}
\paragraph{Data Parallelism (DP).} DP divides the training examples among multiple workers~\cite{li2014scaling,ho2013more} and replicates the model at each worker. During each iteration, each worker calculates the model gradient based on its assigned examples and then synchronizes the gradient with the other workers~\cite{sergeev2018horovod}. However, DP requires each worker to compute and synchronize gradients for the entire model, which can become challenging as the model size increases. One issue is that the large gradients can create a communication bottleneck, and several previous studies have proposed gradient compression methods~\cite{seide20141, bernstein2018signsgd,dettmers20158, lin2017deep, wang2018atomo} to address this. Additionally, the worker may not have enough memory to train with the entire model using even one example, in which case model parallelism may be necessary. 
\paragraph{Model Parallelism (MP).} Model parallelism (MP) divides the model among multiple workers, allowing large models to be trained by only requiring each worker to maintain a portion of the entire model in memory. There are two main paradigms for MP: inter-layer pipeline model parallelism (PP) and intra-layer tensor model parallelism (TP). PP divides the layers among workers, with each worker executing the forward and backward computations in a pipelined fashion across different training examples~\cite{narayanan2019pipedream, li2021terapipe}. For example, a mini-batch of training examples can be partitioned into smaller micro-batches~\cite{huang2019gpipe}, with the forward computation of the first micro-batch taking place on one worker while the forward computation of the second micro-batch happens on another worker in parallel. TP~\cite{lu2017flexflow,shazeer2018mesh,kim2016strads} divides the tensor computations among workers. In particular, we consider a specialized strategy developed for Transformer models that divides the two GEMM layers in the attention module column-wise and then row-wise, with the same partitioning applied to the MLP module~\cite{shoeybi2019megatron}. However, TP still involves a communication bottleneck due to the need for two all-to-all collective operations in each layer, motivating the use of compression to reduce the communication overhead of MP~\cite{shoeybi2019megatron}.
two all-to-all collective operations in each layer~\cite{shoeybi2019megatron}. This bottleneck motivates our study to use compression for reducing the communication of model parallelism.

\subsection{Challenges in Model Parallelism Compression}
\label{sec:challenge}
In data parallelism, synchronizing gradients in large models is a major bottleneck, and several gradient compression algorithms have been proposed~\cite{seide20141, bernstein2018signsgd,dettmers20158, lin2017deep, wang2018atomo} to reduce the communication volume. These algorithms often rely on the observation that the gradient matrix is low-rank. In model parallelism, we have observed that communicating activations becomes the bottleneck. However, we have identified three challenges when adapting gradient compression algorithms for use in model parallelism.

First, the low-rank observation for gradient matrices does not hold for activation matrices, as shown in Figure~\ref{fig:low_rank}. The sigma value percentage for activation matrices increases nearly linearly with the dimension percentage, indicating that the activation matrix is not low-rank. Therefore, applying gradient compression techniques to activations is likely to result in a significant loss of accuracy. Second, the performance of compression methods is heavily influenced by system optimizations~\cite{li2020pytorch}, and many gradient compression methods do not lead to speed-ups for data parallelism~\cite{zhang2017poseidon, agarwal2022utility} due to competition for GPU resources between gradient encoding computation and backward computation. However, the impact of these optimizations on compression methods in model parallelism has not been studied. Third, model parallelism introduces the possibility of using learning-based compression methods, such as autoencoders (AE)~\cite{hinton1993autoencoders}, which have not been examined in the gradient compression literature because they require gradient computations and raise new considerations. Given these three challenges, there is a need for a thorough study of the effects of different compression methods in model parallelism.

\section{Implementation}
In this section, we first introduce the compression algorithms we evaluate in this work (\S~\ref{sec:compress_alg}). Then, we discuss implementation details in Sections~\ref{sec:tensor_parallel_compress} and~\ref{sec:pipeline_parallel_compress}.


\subsection{Compression Algorithms}
\label{sec:compress_alg}
\begin{figure*}[!ht]
    \centering
    \includegraphics[width=\linewidth]{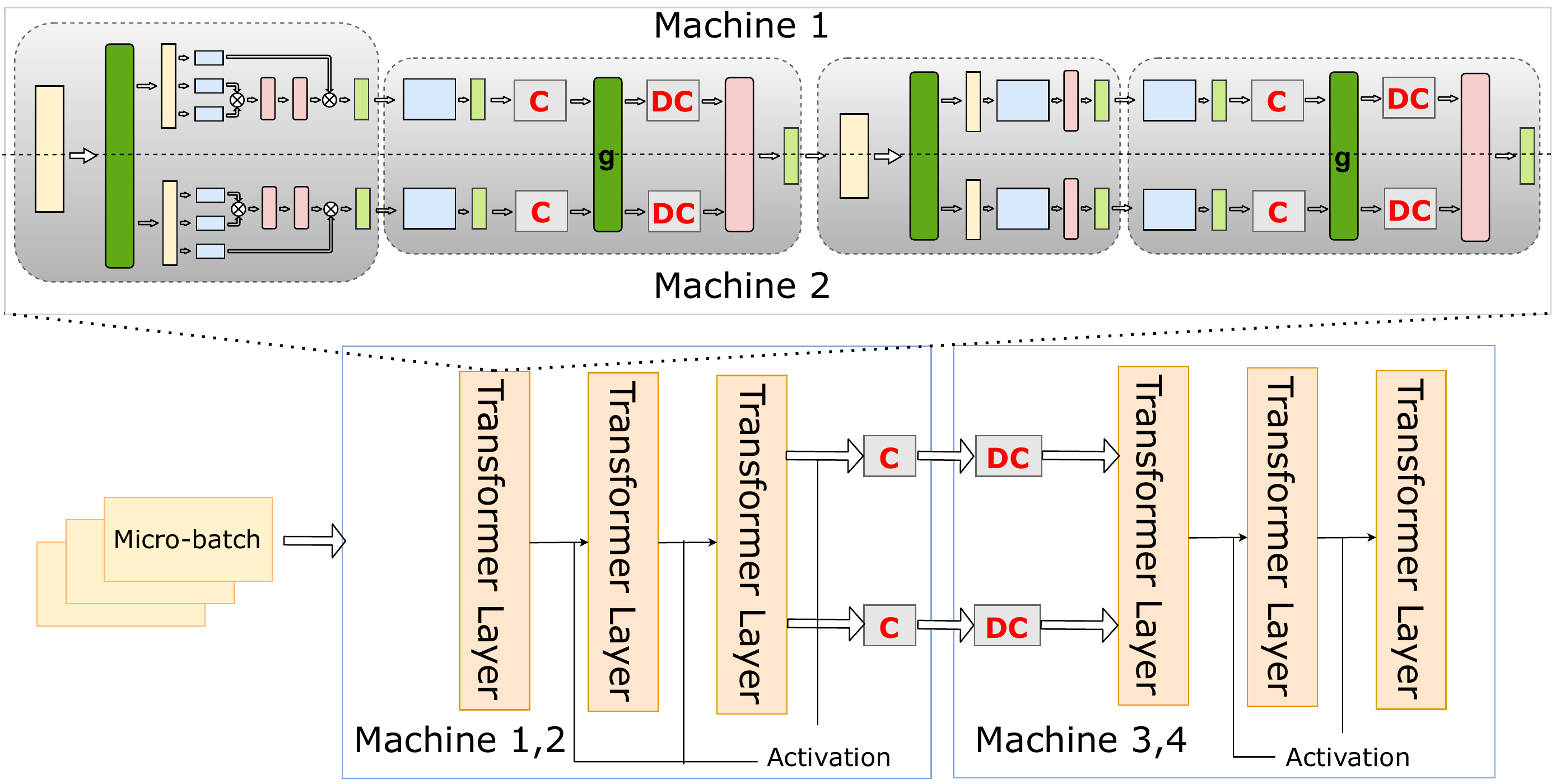}
    \vspace{-6mm}
    \caption{Illustration of compression on a 6-Layer Transformer model with 4 machines. Machine 1 and Machine 2 maintain the first three layers according to the TP strategy (pipeline stage 1). g stands for an all-reduce operation in the forward pass. A compression method C is used to reduce the message size for the all-reduce operation to reduce TP communication time. Correspondingly, a de-compression method DC is used after the communication. For instance, if AE are used, then C is an encoder, and DC is a decoder. Machine 3 and Machine 4 are responsible for the last three layers (pipeline stage 2). A compression method is used before Machine 1 sends the activation to Machine 3, and before Machine 2 sends the activation to Machine 4 to reduce PP communication time. The goal of this paper is to study the effect of different pairs of C and DC.}
    \label{fig:mp_pp_compression}
    \vspace{-2mm}
\end{figure*}

In this work, we evaluate a range of popular compression algorithms, including sparsification-based approaches, learning-based approaches, and quantization-based approaches (as illustrated in Figure~\ref{fig:mp_pp_compression}). We use Top-$K$ and Random-$K$ as sparsification-based approaches, as they have been well-studied in gradient compression~\cite{stich2018sparsified}. We also implement AEs, which compress messages using a small neural network~\cite{hinton1993autoencoders}. For quantization, we use the same scheme as in previous research~\cite{wang2022fine}, but compare its performance to other compression algorithms in the context of model parallelism, as the prior work only considered pipeline compression over slow networks. Since the activation matrices for models are not low-rank (as shown in Figure~\ref{fig:low_rank}), low-rank based compression algorithms (such as PowerSGD~\cite{vogels2019powersgd}) are not suitable for model parallelism compression, and we do not evaluate any low-rank based compression algorithms in this work.

\subsection{Tensor Parallelism Compression}
\label{sec:tensor_parallel_compress}
We base our implementation on Megatron-LM~\cite{shoeybi2019megatron}, a popular Transformer models training system that supports tensor and pipeline model parallelism.
To integrate the compression algorithms into Megatron-LM, we make the following modifications. For AE, we compress the activation before the all-reduce step and invoke the all-reduce function as usual.
The implementation of AE is shown here: 
for each layer, we have a learnable matrix $w \in \R^{h \times c}$, and the activation $X \in \R^{b \times s \times h}$, where $b$ is the batch size, $s$ is the sequence length, $h$ is the hidden size, and $c < h$ is the compressed size. By using the matrix $w$, we output the compressed activation $Xw \in \R^{b \times s \times c}$. Then, we use a similar technique(a decoder as opposed to an encoder) to decompress the compressed activation and propagate it to the next layer. However, since the Top-$K$, Random-$K$, and quantization can output two independent tensors with different types (e.g., for Top-$K$ values and their indices), we cannot use \textsf{torch.distributed.all-reduce} to sum the tensors up directly.
In light of this, we replace the all-reduce step with the all-gather function: \texttt{gather-from-tensor-model-parallel-region}, which is implemented by Megatron-LM. We use \texttt{torch.topk} function to select the $k$ largest absolute values of the activation and \textsf{random.sample} function to randomly select $k$ values from the activation. Finally, our implementation of quantization is based on code released by~\cite{wang2022fine}.

\subsection{Pipeline Parallelism Compression}
\label{sec:pipeline_parallel_compress}
Megatron-LM can only send one tensor to the next pipeline stage per round, so we modify its communication functions to allow for the transmission of multiple tensors per round in order to integrate Top-$K$, Random-$K$, and quantization. Since we compress the activation in the forward step, using compression also reduces the size of the gradient for activation and thus the communication cost in the backward step. However, this is not the case when using quantization to compress the activation for models. This is because, as previously noted~\cite{wang2022fine}, the Pytorch backward engine only supports gradients for floating point tensors, and therefore the size of the gradient is the same as the size of the decompressed activation. Our implementation also allows the integration of error-feedback compression algorithms by retaining the error information from the previous compression step.
\section{experiments}

We next perform experiments using our implementation to answer the following questions:
\begin{itemize}
    \item What is the impact of activation compression on system throughput and which compression method achieves the best throughput? 
    \item What is the impact on the model's accuracy?
    \item How different network bandwidths affect the best compression method?
    \item How do hyper-parameters such as the batch size and sequence length affect the benefits of compression?

\end{itemize}

We answer these questions in the context of two commonly used scenarios in language modeling: fine-tuning on the GLUE benchmark~\cite{wang2018glue}, and pre-training on the Wikipedia~\cite{devlin2018bert} dataset and the BooksCorpus~\cite{zhu2015aligning} dataset. 


\subsection{Experimental Setup}
In this section, we briefly describe the hardware, model, and other experiment settings.

\textbf{System Configuration.} To measure the performance of compression algorithms over different hardware, our experiments are conducted on two different setups. Our first setup uses AWS p3.8xlarge machines which have 4 Tesla V100 GPUs with all GPUs connected by NVLink. AWS p3.8xlarge instances have 10 Gbps network bandwidth across instances. Our second setup uses a local machine which also has 4 Tesla V100 GPUs but does not have NVLink. All the GPUs  are connected by a single PCIe bridge. The local server runs Ubuntu 18.04 LTS and the server has 125GB of memory.

\textbf{Model.} We use the $\text{BERT}_\text{LARGE}$ model provided by Megatron-LM~\cite{shoeybi2019megatron} which has 345M parameters. We configure the model to have 24 layers with each layer having a hidden size of 1024 and 16 attention heads. We use fp16 training to train the $\text{BERT}_\text{LARGE}$ model. 

\textbf{Experimental Settings.} 
For fine-tuning, we follow the previous work~\cite{devlin2018bert, liu2019roberta}, and use micro-batch size 32 and sequence length 512 by default.
We use one machine with 4 V100 GPUs and vary the tensor model-parallel size and the pipeline model-parallel size across the following three parallelism degrees: $\{(1, 4), (2, 2), (4, 1)\}$, where the first number of the tuple represents the tensor model-parallel degree and the second number of the tuple stands for the pipeline model-parallel degree. To investigate the impact of hyper-parameters, we conduct experiments that vary the batch size from $\{8, 32\}$, and sequence length from $\{128, 512\}$ on fine-tuning.


For pre-training, we use micro-batch size 128, global batch size 1024, and sequence length 128. To study the impact of the distributed settings, we use the following three different parallelism degrees: $\{(2, 8), (4, 4), (8, 2)\}$, where the first number of the tuple represents the tensor model-parallel degree and the second number of the tuple represents the pipeline model-parallel degree.

We also evaluate compression algorithms with different parameters. 
For AE, we use different dimension after compression from $\{50, 100\}$. For Top-$K$ and Random-$K$ algorithms, we use two comparable settings: (1) Keep the same compression \textit{ratio} as AE (i.e., we compress the activation around 10 and 20 times.) (2) Keep the same communication \textit{cost} as AE. Finally, we also tune the parameters for quantization and compress the activation to $\{2, 4, 8\}$ bits. 

By default, we perform experiments on $\text{BERT}_{\text{Large}}$ model with 24 layers and compress the activation for the last 12 layers. For instance, when the pipeline model-parallel degree is 2 and tensor model-parallel degree is 2, we compress the activation between two pipeline stages and the communication cost over tensor parallelism in the last 12 layers. We also vary the number of layers compressed in Section~\ref{sec:impact_times_location}.
\begin{table}[!ht]
\centering
{\small
\begin{tabular}{cc}
\rowcolor{Gray} \toprule
 Notation & Description \bigstrut\\
\midrule
\rowcolor{LightCyan} A1 & AE with encoder output dimension 50 \bigstrut\\ 
A2 & AE with encoder output dimension 100 \bigstrut\\
\midrule
\rowcolor{LightCyan} T1/R1 & Top/Rand-$K$: same comm. cost as A1 \bigstrut\\ 
T2/R2 & Top/Rand-$K$: same comm. cost as A2 \bigstrut\\ 
\rowcolor{LightCyan} T3/R3 & Top/Rand-$K$: same comp. ratio as A1 \\ 
T4/R4 & Top/Rand-$K$: same comp. ratio as A2 \bigstrut\\ 
\midrule
\rowcolor{LightCyan} Q1 & Quantization: reduce the precision to 2 bits \\ 
Q2 & Quantization: reduce the precision to 4 bits \bigstrut\\ 
\midrule
\rowcolor{LightCyan} TP & Tensor model-parallelism degree \bigstrut\\
PP & Pipeline model-parallelism degree \bigstrut\\
\bottomrule
\end{tabular}}
\caption{Notation Table. For ease of notation, we use TP/PP to denote the degree of tensor/pipeline model parallelism. `comm' and `comp' are short for `communication' and `compression'.
}
\label{tab:notation}
\vspace{-5mm}
\end{table}

\begin{table*}[!ht]
\centering
{\small
\begin{tabular}{c c c c c c c c}
\rowcolor{Gray} \toprule
Distributed Setting & w/o & A1 & A2 & T1 & T2 & T3 & T4 \bigstrut\\ 
\midrule
\rowcolor{LightCyan} TP=1, PP=4 & 591.96 & \underline{591.36} & \underline{591.47} & 594.81 & 595.53 & 599.65 & 605.05 \bigstrut\\ 
TP=2, PP=2 & 440.71 & \underline{437.98} & 444.02 & 465.73 & 473.64 & 493.16 & 528.93  \bigstrut\\ 
\rowcolor{LightCyan} TP=4, PP=1 & \textbf{261.48} & 270.22 & 275.54 & 314.37 & 323.90 & 356.57 & 409.23  \bigstrut\\ 
\bottomrule
\rowcolor{Gray} Distributed Setting & w/o & R1 & R2 & R3 & R4 & Q1 & Q2 \bigstrut\\
\midrule
\rowcolor{LightCyan} TP=1, PP=4 & 591.96 & 749.56 & 1,008.64 & 1,824.36 & 5,572.87 & 595.29 & 595.45 \bigstrut\\
TP=2, PP=2 & 440.71 & 3,377.59 & 6,616.30 & 17,117.01 & 71,058.64 & 489.27 & 486.54 \bigstrut\\ 
\rowcolor{LightCyan} TP=4, PP=1 & \textbf{261.48} & 3,254.01 & 6,561.22 & 16,990.88 & 65,121.79 & 347.68 & 350.50 \bigstrut\\
\bottomrule
\end{tabular}}
\caption{The average iteration time (ms) for fine-tuning with various compression techniques by varying the distributed setting. The results are collected from the AWS p3.8xlarge machine with NVLink by using batch size 32, and sequence length 512. The best setting is \textbf{bolded} in the table. And the settings which see benefits compared with the baseline, are \underline{underlined}.}
\label{tab:run_time_finetune_AWS}
\end{table*}

\begin{table}[!ht]
\centering
{\small
\begin{tabular}{c c c c}
\rowcolor{Gray} \toprule
With NVLink & w/o & A1 & A2 \bigstrut\\
\midrule
\rowcolor{LightCyan} TP=1, PP=4 & 591.96 & \underline{591.36} & \underline{591.47} \bigstrut\\
TP=2, PP=2 & 440.71 & \underline{437.98} & 444.02 \bigstrut\\
\rowcolor{LightCyan} TP=4, PP=1 & \textbf{261.48} & 270.22 & 275.54 \bigstrut\\
\bottomrule
\rowcolor{Gray} Without NVLink & w/o & A1 & A2 \bigstrut\\
\midrule
\rowcolor{LightCyan} TP=1, PP=4 & 633.17 & \underline{620.10} & \underline{620.44} \bigstrut\\
TP=2, PP=2 & 646.14 & \textbf{\underline{586.65}} & \underline{595.25} \bigstrut\\
\rowcolor{LightCyan} TP=4, PP=1 & 736.01 & \underline{624.62} & \underline{636.15} \bigstrut\\
\bottomrule
\end{tabular}}
\caption{The average iteration time (ms) for fine-tuning with/without NVLink.
We compare time without compression and with AE on different distributed settings, with batch size 32, and sequence length 512. The best setting on each machine is \textbf{bolded}. And the settings, under which we can gain benefits compared with the baseline, are \underline{underlined}.}
\label{tab:run_time_finetune_local}
\vspace{-8mm}
\end{table}

\begin{table*}[!ht]
\centering
\scalebox{0.95}{
{\small
\begin{tabular}{c c c c c c c c c c c}
\rowcolor{Gray} \toprule \makecell[c]{Compression \\ Algorithm} & Forward & Backward & Optimizer & \makecell[c]{Waiting \& \\ Pipeline Comm.} & Total Time & Tensor Enc. & Tensor Dec. & \makecell[c]{Tensor \\ Comm.} \bigstrut\\
\midrule
\rowcolor{LightCyan} w/o & 276.34 & 354.16 & 5.80 & 9.83 & 646.14 &  $\backslash$ & $\backslash$ & 150.72 \bigstrut\\
\midrule
A1 & 213.83 & 362.61 & 6.16 & 4.06 & 586.65 & 2.16 & 3.12 & 80.88 \bigstrut\\
\rowcolor{LightCyan} A2 & 219.01 & 366.51 & 5.67 & 4.07 & 595.25 & 3.12 & 4.56 & 84.48 \bigstrut\\
\midrule
T1 & 298.93 & 355.71 & 6.79 & 4.38 & 665.81 & 70.08 & 13.68 & 85.20 \bigstrut\\
\rowcolor{LightCyan} T2 & 305.47 & 355.51 & 6.36 & 3.91 & 671.24 & 70.32 & 16.80 & 87.84 \bigstrut\\
T3 & 331.70 & 356.80 & 5.78 & 5.00 & 699.27 & 72.24 & 27.36 & 100.80 \bigstrut\\
\rowcolor{LightCyan} T4 & 376.72 & 359.19 & 5.89 & 6.60 & 748.41 & 74.88 & 45.36 & 124.56 \bigstrut\\
\midrule
R1 & 2,408.68 & 357.02 & 6.10 & 7.68 & 2,779.49  & 2,040.24 & 15.84 & 104.16  \bigstrut\\
\rowcolor{LightCyan} R2 & 4,696.99 & 356.33 & 6.28 & 6.20 & 5,065.80  & 4,244.64 & 19.44 & 135.84 \bigstrut\\
R3 & 12,603.79 & 362.13 & 6.81 & 25.28 & 12,998.01 & 11,499.12 & 29.76 & 139.92 \\
\rowcolor{LightCyan} R4 & 46,968.21 & 365.36 & 7.61 & 22.81 & 47,363.98 & 44,038.56 & 47.52 & 567.36 \bigstrut\\
\midrule
Q1 & 274.03 & 354.56 & 5.88 & 7.98 & 642.46 & 20.64 & 32.16 & 91.68 \bigstrut\\
\rowcolor{LightCyan} Q2 & 282.64 & 354.55 & 5.58 & 7.58 & 650.36 & 19.92 & 30.24 & 104.64 \bigstrut\\
\bottomrule
\end{tabular}}}
\vspace{-1mm}
\caption{We breakdown the average iteration time (ms) for fine-tuning with various compression techniques when using TP=2 and PP=2, batch size 32, and sequence length 512. The results are collected from the local machine without NVLink. The total time (ms) is divided into following parts: forward step, backward step, optimizer, and waiting \& pipeline communication. The last three columns further breakdown the tensor encoder/decoder and communication times which are considered part of the forward step.}
\label{tab:breakdown_run_time_finetune_local}
\end{table*}

\begin{table*}[!ht]
\centering
{\small
\begin{tabular}{c c c c c c c c c c}
\rowcolor{Gray} \toprule
\makecell[c]{Compression \\ Algorithm} & MNLI-(m/mm) & QQP & SST-2 & MRPC & CoLA & QNLI & RTE & STS-B & Avg.\bigstrut\\
\midrule
\rowcolor{LightCyan} w/o & 88.07/88.70 & 92.02 & 95.07 & 88.46 & 62.22 & 93.39 & 82.67 & 89.16 & 86.64 \bigstrut\\
\midrule
 A1 & 85.42/85.43 & 91.07 & 92.09 & 86.14 & 54.18 & 91.31 & 70.04 & 87.61 & 82.59  \bigstrut\\
\rowcolor{LightCyan} A2 & 85.53/85.65 & 91.24 & 93.23 & 85.86 & 55.93 & 91.01 & 65.34 & 87.76 & 82.40 \bigstrut\\
\midrule
 T1 & 32.05/32.18 & 74.31 & 83.60 & 70.78 & 0.00 & 58.37 & 51.99 & 0.00 & 44.81  \bigstrut\\
\rowcolor{LightCyan} T2 & 44.12/45.67 & 39.68 & 90.83 & 78.09 & 0.00 & 84.42 & 49.82 & 62.70 & 55.04 \bigstrut\\
 T3 & 36.12/36.08 & 74.75 & 90.25 & 81.51 & 0.00 & 85.41 & 54.15 & 0.00 & 50.92 \bigstrut\\
\rowcolor{LightCyan} T4 & 83.85/84.41 & 56.39 & 93.69 & 83.65 & 0.00 & 90.54 & 59.21 & 86.02 & 70.86 \bigstrut\\
\midrule
 Q1 & 87.25/87.81 & 91.71 & 93.46 & 87.01 & 55.99 & 61.38 & 67.51 & 88.02 & 80.02  \bigstrut\\
\rowcolor{LightCyan} Q2 & 87.85/88.47 & 91.93 & 93.23 & 87.42 & 57.67 & 93.01 & 78.34 & 87.43 & 85.04 \bigstrut\\
\bottomrule
\end{tabular}}
\vspace{-1mm}
\caption{Fine-tuning results over GLUE dataset under the setting that the tensor model-parallel size is 2 and pipeline model-parallel size is 2. F1 scores are reported for QQP and MRPC, Matthews correlation coefficients are reported for CoLA, and Spearman correlations are reported for STS-B, and accuracy scores are reported for the other tasks.}
\label{tab:accuracy_finetune}
\end{table*}

\begin{table*}[!ht]
\centering
{\small
\begin{tabular}{c c c c c c c c}
\toprule
\rowcolor{Gray} Distributed Setting & w/o & A1 & A2 & T1 & T2 & T3 & T4 \bigstrut\\
\midrule
\rowcolor{LightCyan} TP=2, PP=8 & 1,625.16 & \underline{1,550.18} & \underline{1,579.70} & \underline{1,508.34} & \underline{1,503.54} & \underline{1,593.37} & 1,682.87 \bigstrut\\
TP=4, PP=4 & 1,422.40 & \underline{1,242.97} & \underline{\textbf{1,223.20}} & \underline{1,360.37} & \underline{1,352.61} & \underline{1,410.47} & 1,721.87 \bigstrut\\
\rowcolor{LightCyan} TP=8, PP=2 & 15,642.30 &
\underline{14,577.29} & \underline{14,073.45} & \underline{14,308.12} & \underline{14,543.81} & 18,919.92 & 27,152.07 \bigstrut\\
\bottomrule
\rowcolor{Gray} Distributed Setting & w/o & R1 & R2 & R3 & R4 & Q1 & Q2 \bigstrut\\
\midrule
\rowcolor{LightCyan} TP=2, PP=8 & 1,625.16 & 10,308.03 & 20,814.20 & 55,925.28 & $>$100,000 & 1,759.27 & 1,752.24 \bigstrut\\
TP=4, PP=4 & 1,422.40 & 15,433.12 & 31,565.19 & 87,421.46 & $>$100,000 & 2,435.03 & 2,594.94 \bigstrut\\
\rowcolor{LightCyan} TP=8, PP=2 & 15,642.30 & 32,522.47 & 61,049.87 & $>$100,000 & $>$100,000 & 16,414.57 & 16,517.44 \bigstrut\\
\bottomrule
\end{tabular}}
\vspace{-1mm}
\caption{The average iteration time (ms) for pre-training with various compression techniques by varying the distributed setting. The results are collected from 4 AWS p3.8xlarge machines with NVLink by using micro-batch size 128, global batch size 1024, and sequence length 128. The best setting is \textbf{bolded} in the table. And the settings, under which we can gain benefits compared with the baseline, are \underline{underlined}.}
\label{tab:run_time_pretrain}
\end{table*}

\begin{table*}[!ht]
\centering
\scalebox{0.95}{{\small
\begin{tabular}{c c c c c c c c c}
\rowcolor{Gray} \toprule \makecell[c]{Compression \\ Algorithm} & Forward & Backward & Optimizer & \makecell[c]{Waiting \& \\ Pipeline Comm.} & Total Time & Tensor Enc. & Tensor Dec. & \makecell[c]{Tensor \\ Comm.} \bigstrut\\
\midrule
\rowcolor{LightCyan} w/o & 467.73 & 419.26 & 7.42 & 527.99 & 1,422.40 & $\backslash$ & $\backslash$ & 91.08 \bigstrut\\
\midrule
A1 & 546.95 & 455.26 & 7.29 & 233.47 & 1,242.97 & 8.64 & 16.20 & 32.76 \\
\rowcolor{LightCyan} A2 & 459.26 & 467.51 & 9.64 & 286.78 & 1,223.20 & 12.96 & 20.52 & 43.56 \bigstrut\\
\midrule
T1 & 712.22 & 423.91 & 7.21 & 217.03 & 1,360.37 & 73.44 & 140.4 & 80.28 \bigstrut\\
\rowcolor{LightCyan} T2 & 671.19 & 424.27 & 7.35 & 249.80 & 1,352.61 & 81.00 & 170.64 & 81.36 \bigstrut\\
T3 & 813.03 & 433.42 & 7.35 & 156.67 & 1,410.47 & 108.00 & 268.92 & 115.92 \bigstrut\\
\rowcolor{LightCyan} T4 & 1,068.38 & 444.26 & 6.75 & 202.48 & 1,721.87 & 153.36 & 427.68 & 151.56 \bigstrut\\
\midrule
R1 & 14,199.56 & 421.40 & 4.23 & 807.93 & 15,433.12 & 13,185.72 & 181.44 & 193.68 \\
\rowcolor{LightCyan} R2 & 29,344.85 & 427.18 & 3.91 & 1,789.25 & 31,565.19 & 27,975.24 & 181.44 & 187.20 \bigstrut\\
R3 & 78,906.91 & 444.88 & 6.08 & 3,707.37 & 83,065.23 & 73,847.16 & 279.72 & 649.44 \bigstrut\\
\midrule
\rowcolor{LightCyan} Q1 & 803.63 & 417.33 & 8.61 & 1,205.46 & 2,435.03 & 90.72 & 304.56 & 193.68 \bigstrut\\
Q2 & 805.33 & 417.74 & 7.55 & 1,364.32 & 2,594.94 & 85.32 & 271.08 & 111.60 \bigstrut\\
\bottomrule
\end{tabular}}}
\vspace{-1mm}
\caption{We breakdown the average iteration time (ms) for pre-training with various compression techniques when using tensor model-parallel size 4, pipeline model-parallel size 4, micro batch size 128, global batch size 1024, and sequence length 128. The results are collected from 4 AWS p3.8xlarge machines with NVLink.} 
\label{tab:breakdown_run_time_pretrain}
\end{table*}

\begin{table*}[!ht]
\centering
{\small
\begin{tabular}{c c c c c c c c c c}
\toprule
\rowcolor{Gray} \makecell[c]{Compression \\ Algorithm} & MNLI-(m/mm) & QQP & SST-2 & MRPC & CoLA & QNLI & RTE & STS-B & Avg. \bigstrut\\ 
\midrule
\rowcolor{LightCyan} w/o & 84.87/84.79 & 91.25 & 92.43 & 86.84 & 56.36 & 92.26 & 70.40 & 86.83 & 82.89 \bigstrut\\
A2 & 83.77/84.32 & 91.14 & 91.63 & 86.55 & 58.61 & 91.96 & 71.48 & 87.16 & 82.96 \bigstrut\\
\rowcolor{LightCyan} T2 & 61.06/60.93 & 80.74 & 80.16 & 63.83 & 10.01 & 59.55 & 47.29 & 0.37 & 51.55 \bigstrut\\ 
Q2 & 84.47/85.32 & 91.36 & 93.23 & 85.10 & 58.84 & 91.69 & 71.84 & 86.39 & 83.14 \bigstrut\\
\bottomrule
\end{tabular}}
\vspace{-1mm}
\caption{Fine-tuning results over GLUE dataset by using the checkpoint obtained by pre-training. F1 scores are reported for QQP and MRPC, Matthews correlation coefficient is reported for CoLA, and Spearman correlations are reported for STS-B, and accuracy scores are reported for the other tasks.}
\label{tab:accuracy_finetune_pretrain}
\end{table*}

\subsection{Throughput Benefits for Fine-Tuning}
\label{sec:throughput_ft}
\begin{takeaway}
Using non-learning-based compression techniques to compress activations only slightly improves system throughput (by 1\% or less) due to the large overhead of these methods. However, we see end-to-end speedups of up to 17.8\% when using learning-based compression methods on a machine without NVLink.
\end{takeaway}
When running fine-tune experiments on a p3.8xlarge instance on Amazon EC2, we cannot improve system throughput by using non-learning-based compression algorithms (Table~\ref{tab:run_time_finetune_AWS}). Comparing  Tables~\ref{tab:run_time_finetune_AWS} and~\ref{tab:run_time_finetune_local}, we can see that the network bandwidth across the GPUs can affect the performance benefits from compression. In other words, we can improve system throughput by at most 17.8\% 
when compressing activation for fine-tuning tasks on a 4-GPU machine without NVLink. That's because, without NVLink, the communication time for model parallelism is much longer. Thus, while the message encoding and decoding time remain unchanged, compression methods can provide more throughput benefits across lower bandwidth links.

Furthermore, from Tables~\ref{tab:run_time_finetune_AWS} and \ref{tab:run_time_finetune_local}, we observe that AE outperforms other compression methods. In Table~\ref{tab:breakdown_run_time_finetune_local}, we breakdown the time taken by each algorithm and find that Top-$K$, Random-$K$ and quantization have large encoding/decoding overheads and thus cannot provide end-to-end throughput improvements. Although AE slightly increases the time taken by the backward step, the $\sim2\times$ reduction in communication time and the limited encoding/decoding overhead lead to better overall throughput.

\subsection{Effect of Compression on Model Accuracy while Fine-tuning}
\label{sec:accuracy_ft}
\begin{takeaway}
Among all evaluated compression algorithms, only AE
and quantization preserve fine-tuning accuracy.
\end{takeaway}
From Table~\ref{tab:accuracy_finetune}, we can see that, when using AE and quantization algorithm for compression, the accuracy loss is within 3\% except for CoLA and RTE. In Figure~\ref{fig:low_rank}, we have shown that the activation for models is not low-rank. Therefore, sparsification-based compression algorithms (Top-$K$/Random-$K$) lose important information and do not preserve model accuracy. Given that there is significant accuracy difference for CoLA and RTE, we study the impact of varying the number and range of layers compressed for these two datasets in Section~\ref{sec:impact_times_location}.

\subsection{Throughput Benefits for Pre-training}
\label{sec:throughput_pt}
\begin{takeaway}
Only AE and Top-$K$ algorithms improve throughput when performing distributed pre-training. 
\end{takeaway}
First, we recap the experimental environment here. For pre-training, we use 4 p3.8xlarge instances on Amazon EC2 and each instance has 4 GPUs with NVLink. From Table~\ref{tab:run_time_pretrain}, we can see that using Top-$K$ and AE can speed up pre-training by 7\% and 16\% respectively.
Among the three distributed settings, \texttt{TP=4, PP=4} is the best setting for pre-training. That is because the communication cost of tensor parallelism is larger than that of pipeline parallelism and with TP=4, tensor parallel communication happens over faster NVLinks. 


\begin{takeaway}
Compressing activation for models can improve throughput for pre-training by 16\%. 
\end{takeaway}
From Table~\ref{tab:breakdown_run_time_pretrain}, we notice that using AE and Top-$K$ can reduce the waiting time and pipeline communication time of pre-training. This is because the inter-node bandwidth (10Gbps) is smaller than the intra-node bandwidth (40GB/s with NVLink),
so compression is effective at reducing the communication time between two pipeline stages. From Table~\ref{tab:comm_pipeline}, we can observe that, by using A2 to compress the activation over the last 12 layers, we can reduce the communication cost between two pipeline stages effectively.

\begin{table}[!ht]
\centering
{\small
\begin{tabular}{c c c}
\toprule
\rowcolor{Gray} Pipeline Stages & Comm. (w/o) & Comm. (A2) \\ 
\midrule
\rowcolor{LightCyan}$0 \leftrightarrow 1$ & 77.82 & 76.13 \bigstrut\\
$1 \leftrightarrow 2$ & 88.69 & 13.19 \bigstrut\\
\rowcolor{LightCyan}$2 \leftrightarrow 3$ & 97.67 & 14.09 \bigstrut\\
\bottomrule
\end{tabular}}
\vspace{-1mm}
\caption{
The average communication time (ms) per iteration between two pipeline stages. The first column indicates the pipeline stage. And the second column shows the communication time per iteration without compression. Moreover, the third column presents the communication time with A2. We only compress the activation in the last 12 layers and thus the time for the first pipeline stage is unchanged.}
\label{tab:comm_pipeline}
\vspace{-2mm}
\end{table}

\begin{takeaway}
Among all evaluated methods, AE is the best strategy to compress activation over pre-training. It achieves higher pre-training throughput and preserves the model's accuracy.
\end{takeaway}
From Table~\ref{tab:accuracy_finetune_pretrain}, compared with the baseline (without compression), we can observe that using AE is able to keep the accuracy when compared to the uncompressed model. In addition, we observe that we can use the AE at the pre-training phase and remove it during the fine-tuning phase. In other words, we only need to load the parameter of the $\text{BERT}_{\text{Large}}$ model to do fine-tuning, and the parameters of the AE can be ignored. Furthermore, Table~\ref{tab:accuracy_finetune_pretrain} shows that pre-trained models suffer significant accuracy loss when using Top-$K$ for compression. Finally, quantization can preserve the model's accuracy, but we cannot achieve end-to-end speedup by using quantization as strategy to compress activation over pre-training. In conclusion, it is not a good choice to compress the activation by using quantization or Top-$K$.

\subsection{Varying Compression Layers and Location}
\label{sec:impact_times_location}

\begin{figure}[!ht]
\centering
\subfigure[Vary Number of Layers Compressed]{
\includegraphics[scale=0.30]{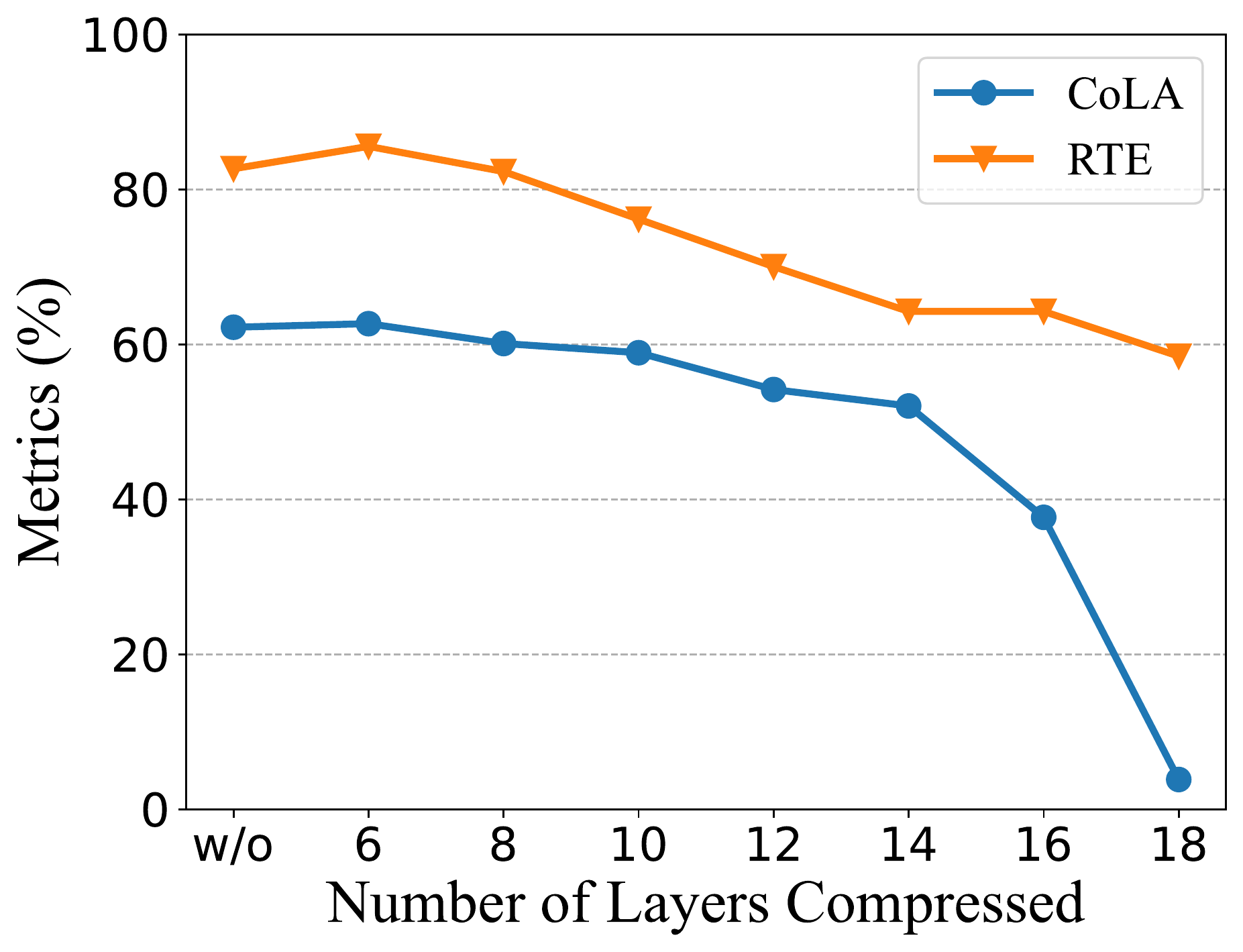}
\label{fig:vary compression times}
}
\subfigure[Vary Compression Location]{  
\includegraphics[scale=0.30]{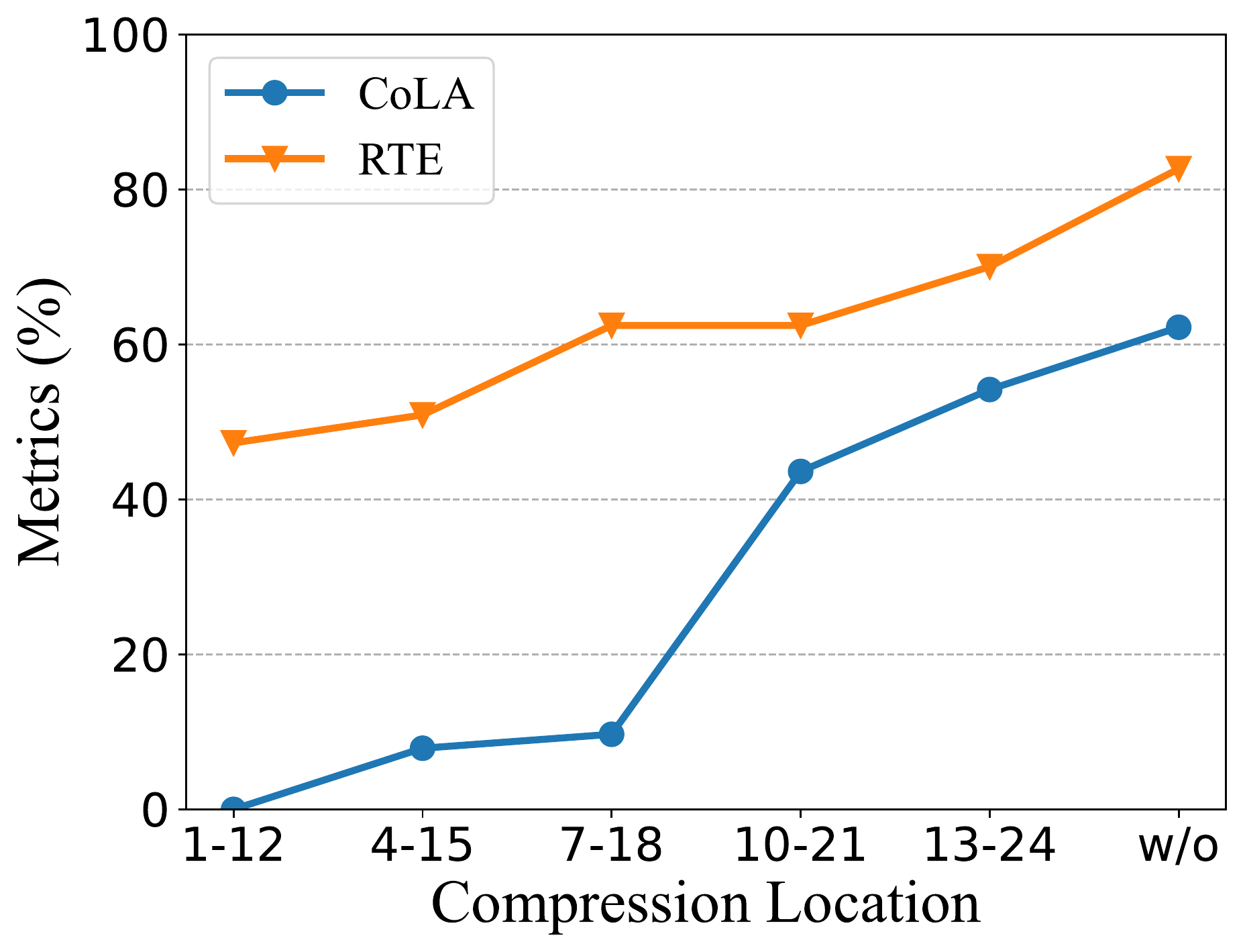} 
\label{fig:vary compression location}
}
\caption{Fine-tuning results over CoLA and RTE datasets by varying the compression location and number of layers compressed.
The above figure shows that model performance vs the number of layers compressed. The below figure shows that model performance versus the compression location.
We use tensor model-parallel degree 2, pipeline model-parallel degree 2, batch size 32, and sequence length 512.}
\label{fig:vary compress times and location}
\end{figure}


\begin{takeaway}
When the number of compressed layers increases, the model accuracy decreases.
\end{takeaway}
From Figure~\ref{fig:vary compression times}, we can observe that the accuracy for RTE and the matthews correlation coefficient for CoLA decreases as we increase the number of layers compressed. This is because as we increase number of layers compressed, we lose more information in the activations leading to a loss in accuracy. From Figure~\ref{fig:vary compression times}, we observe that compressing activations of the last 8 layers is the best strategy to keep the accuracy loss within 3\% for both datasets.
\begin{takeaway}
Compressing the activation for the initial layers harms the accuracy of the model.
\end{takeaway}
\vspace{-2mm}
We keep the number of layers compressed constant and vary the location where we apply compression (Figure~\ref{fig:vary compression location}). The results indicate that compressing activations of the first few layers of the model significantly harms the model's accuracy. This is because compressing activations generates error and the error in the early layers can be accumulated and propagated to later layers.



\subsection{Impact of Model Hyper-parameters}
\label{sec:impact_hyper}
\begin{takeaway}
Using a smaller batch size or sequence length for fine-tuning negates the throughput benefits from compression because of the smaller communication cost.
\end{takeaway}
\vspace{-2mm}
We vary the batch size from $\{8, 32\}$ and sequence length from $\{128, 512\}$, and report the results in Table~\ref{tab:run_time_finetune_AWS_32_128}-\ref{tab:run_time_finetune_local_8_128}. We provide more detailed experimental results in Appendix~\ref{sec:more_exp_res}. We notice that when the communication cost over model parallelism is small, the overhead of the compression methods can become the bottleneck. Therefore, we cannot improve system throughput when using compression algorithms with batch size 8 and sequence length 128. 


\subsection{Performance Analysis}

\begin{figure*}[!ht]
\centering
\subfigure[$T_{comp}$]{\label{fig:cost_comp_without_nvlink}\includegraphics[scale=0.25]{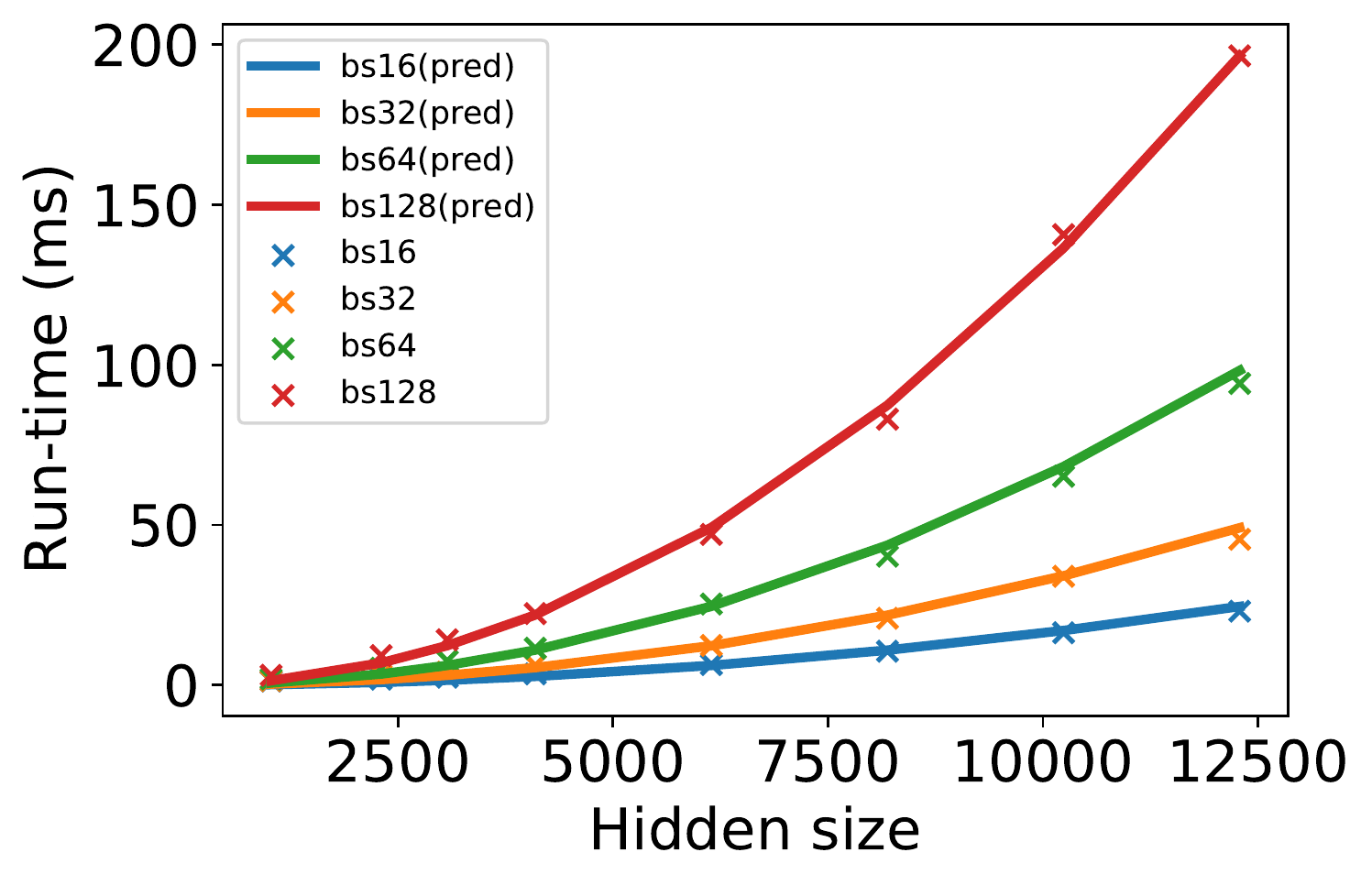}}
\subfigure[$T_{comm}$]{\label{fig:cost_comm_without_nvlink}\includegraphics[scale=0.25]{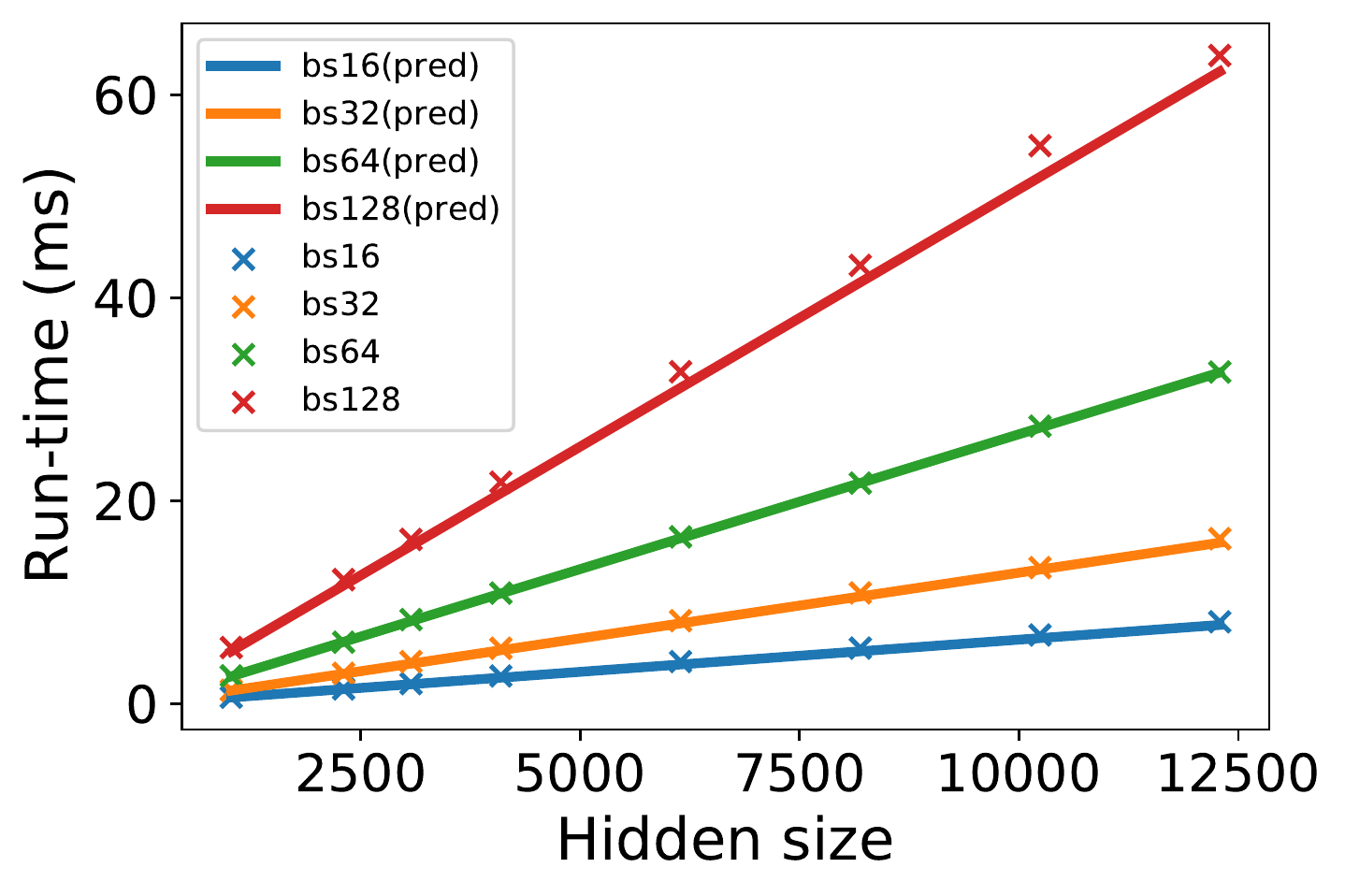}}
\subfigure[$T_{overhead}$]{\label{fig:cost_overhead_without_nvlink}\includegraphics[scale=0.25]{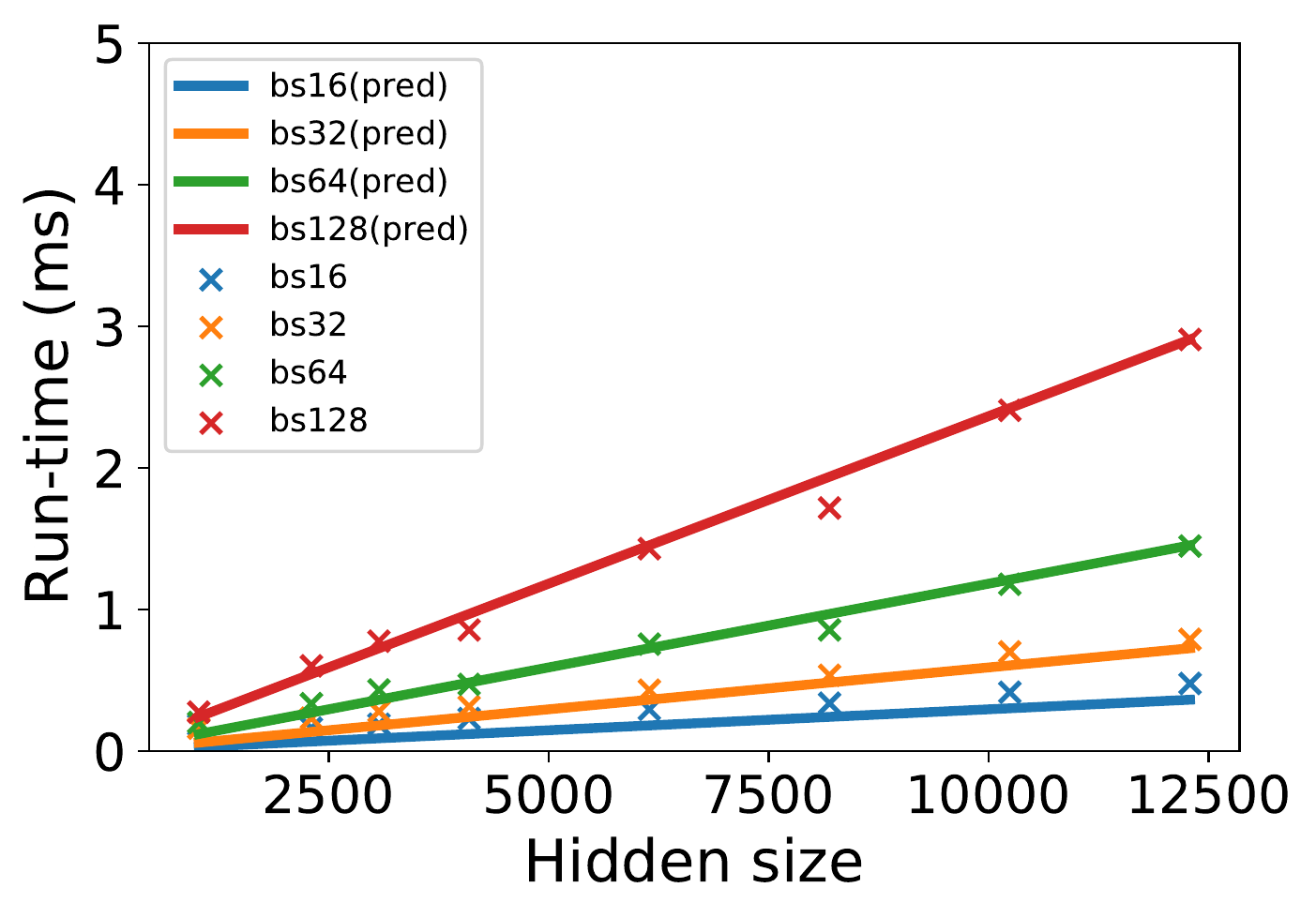}}
\subfigure[Speedup]{\label{fig:overall_speedup_without_nvlink}\includegraphics[scale=0.25]{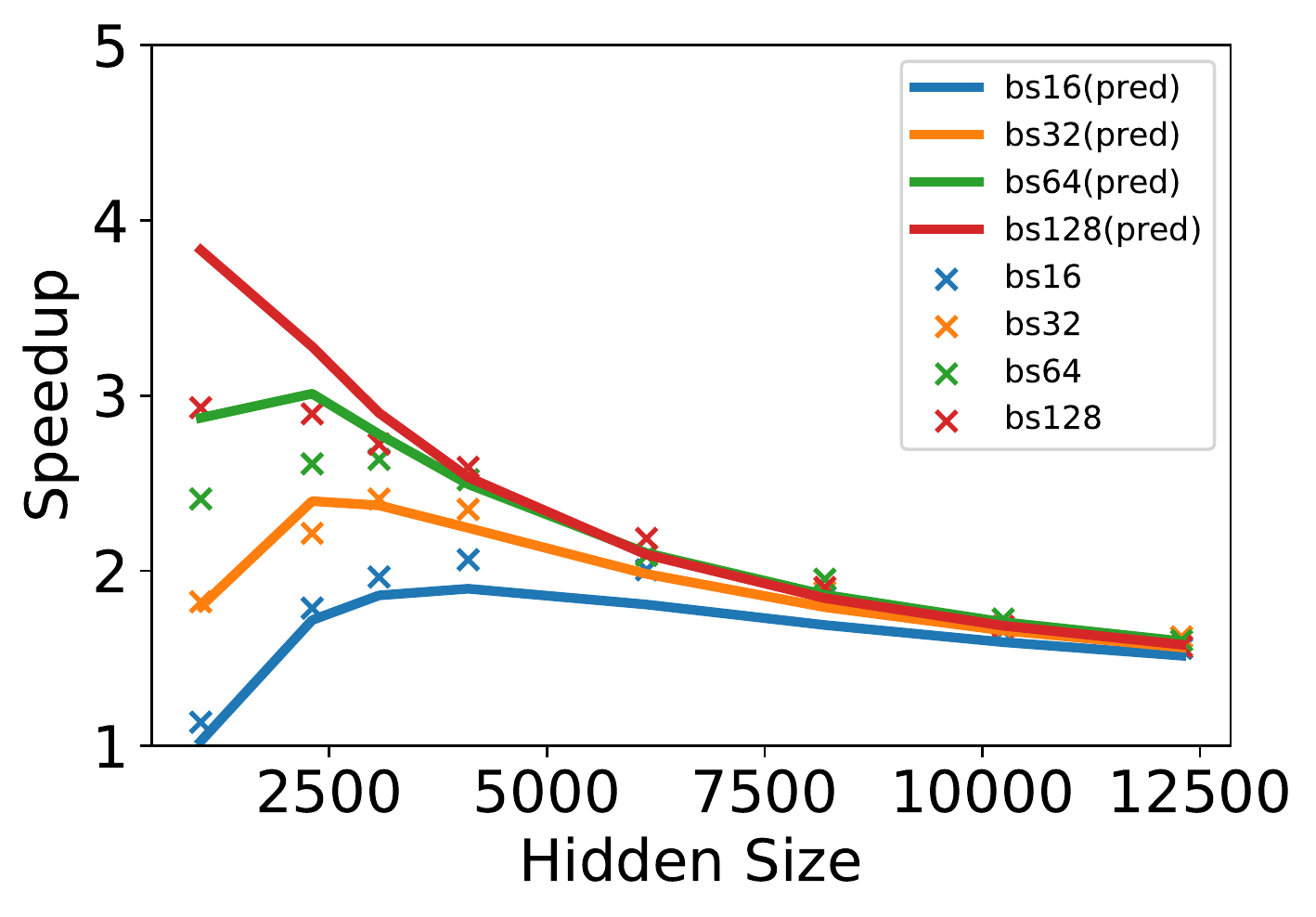}}
\vspace{-4mm}
\caption{We vary the batch size and the hidden size to show that our prediction model is accurate compared with the real experimental results. The model we use here has only one transformer layer and the tensor model-parallel size is 4. In specific, Figure (a) shows the real and predicted computation time with the increase of the hidden size. Figure (b) presents the real and predicted communication time between tensor parallelism by varying the hidden size. As for the Figure (c), it presents the computation time of AE with the increase of hidden size. In the end, Figure (d) show the total speedup when we use AE to compress activations over tensor parallelism.}

\vspace{-2mm}
\end{figure*}

In this section, we develop an analytical cost model to answer the question:
\begin{center}
\textit{What will happen if we scale up the model size and the cluster size?}
\end{center}

Given that prior works~\cite{li2022amp} have analyzed the complexity of various model parallelism strategies, we only consider a fixed strategy of using tensor
model parallelism here. Concretely, we use tensor model parallelism in the same node, and pipeline model parallelism across the node,
a suggested strategy according to~\cite{narayanan2021efficient}. In particular, we build the performance analysis for real-world settings similar to~\cite{narayanan2019pipedream} in two steps. First, we develop our own model on a single-node scenario, and we scale up the model size on a single node. Second, we increase the cluster size and, according to the model-parallelism strategy we choose, assign additional GPUs to pipeline parallelism, and use off-the-shelf pipeline parallelism cost models to predict the performance~\cite{li2022amp,zheng2022alpa}.


Denote the vocabulary size as $V$, hidden size as $h$, sequence length as $s$, and batch size as $B$. From~\cite{narayanan2021efficient}, we know that the number of floating points operations (FLOPs) and all-reduce message size in a Transformer layer is $96Bsh^2 + 16Bs^2h$, and $Bsh$ respectively.

If we do not use compression methods, the total time of a Transformer layer can be modeled as a sum of the all-reduce communication step and the computation time step. We note that these two steps can hardly overlap because
, the reason behind it is that  
the all-reduce communication depends on the previous computational results:
\begin{align}
\label{eq:without_compression}
    T = T_{comp}(96Bsh^2 + 16Bs^2h) + T_{comm}(Bsh)
\end{align} 

\paragraph{Modeling $T_{comp}$} We model $T_{comp}$ as a linear function of FLOPs with the coefficient $\alpha$ that corresponds to the peak performance of the GPU. In particular, we estimate $\alpha$ using ground truth wall clock time of the largest hidden size we can fit, where the GPU is more likely to be of the peak utilization ~\cite{williams2009roofline}. During experiments, we found that fitting $\alpha$ using time of smaller hidden sizes can result in a 30x higher prediction time when we scale up the hidden size because of low GPU utilization. 
Our prediction versus the ground truth time is plotted in Figure~\ref{fig:cost_comp_without_nvlink}. 


\paragraph{Modeling $T_{comm}$} we model $T_{comm}$ as a piece-wise function of the message size~\cite{agarwal2022utility}. Formally, 
\begin{align*}
    T_{comm}(Bsh)=
    \begin{cases}
        c & \text{if } Bsh < d \\
        \beta Bsh & \text{if } Bsh \geq d 
    \end{cases}
\end{align*}
If the message size is smaller than a threshold $d$, then $T_{comm}(Bsh)$ is a constant $c$ because the worker needs to launch one communication round~\cite{li2020pytorch}. Otherwise, the number of communication rounds is proportional to the message size.
The fitting result is in Figure~\ref{fig:cost_comm_without_nvlink}.



Using AE as the compression method and a fixed encoder dimension $e$ (we set $e$ to 100 in this section), the total time of a single Transformer layer is:
\begin{align*}
    T_{AE} &= T_{comp}(96Bsh^2 + 16Bs^2h) + T_{comm}(Bse) \\
    &\quad +T_{overhead}
\end{align*}
Compared with the setting without compression, the computation time remains unchanged. In addition, $T_{comm}(Bse)$ is roughly equal to $c$ because $Bse$ is usually smaller than the threshold $d$. In our experiments, the threshold $d = 16 \times 128 \times 100 = 409600$ and $c \approx 0.2$.

\paragraph{Modeling $T_{overhead}$} In AE, $T_{overhead}$ is the encoder and decoder computation time. It is a batched matrix multiplication with input dimension $B \times s \times h$ and $h \times e$. Assuming $e$ is kept constant, it can be modeled as $T_{overhead} = \gamma Bsh$.
The fitting result is shown in~\ref{fig:cost_overhead_without_nvlink}. 

Since each Transformer layer has identical configurations in popular Transformer models~\cite{devlin2018bert, radford2018improving}, the overall speedup ratio is the same as we vary the number of layers. Thus, we can estimate the speedup of different hidden sizes of any number of Transformer layers using $\frac{T}{T_{AE}}$. We provide the fitting result in Figure~\ref{fig:overall_speedup_without_nvlink}.

\paragraph{Understanding the trend}
We consider the asymptotic behavior of large hidden size h:
\begin{align}
\label{eq:speed_up}
    \frac{T}{T_{AE}} \approx \frac{\alpha (96Bsh^2 + 16 Bs^2h) + \beta Bsh}{\alpha (96Bsh^2 + 16Bs^2h) + \gamma Bsh + c} 
\end{align}
\textbf{Thus, we can see that as hidden layer size increases, the benefits from compression diminish.}


\paragraph{Scaling up the cluster size} Next we analyze the speedup when scaling up the cluster size by combining the pipeline parallelism cost model developed in~\cite{li2022amp,zheng2022alpa}. Formally, the running time is modeled as a sum of per-micro-batch pipeline communication time, per-micro-batch of non-straggler pipeline execution time, and the \textit{per-mini-batch} straggler pipeline execution time. To use the cost model, we denote the micro-batch size as $m$, the number of nodes $n$, the number of layers $L$, the pipeline communication time $p$ or $p_{AE}$.

We use the default pipeline layer assignment strategy in~\cite{shoeybi2019megatron}, which balances the number of transformer layers. Thus, every stage takes the same time in our scenario: $\frac{L}{n}T$ or $\frac{L}{n}T_{AE}$. We use the pipeline communication model in~\cite{jia2019beyond, li2022amp}, $p = \frac{Bsh}{w}$, $p_{AE} = \frac {Bse}{w}$, where $w$ is the bandwidth. Thus the overall speedup can be written as:
\begin{equation}
    \frac{(\frac{m-1}{n}+1) \times LT + (n-1) \times \frac{Bsh}{w}}{(\frac{m-1}{n}+1) \times LT_{AE} + (n-1) \times \frac{Bse}{w}}
\end{equation}
From the Table~\ref{tab:up_cluster}, we see that we can maintain a $\sim$1.5x speedup as we scale the hidden size to 25600. This shows that if we increase the number of nodes when we increase in hidden size, AE compression retains its benefits. However, it is possible to avoid the diminishing speedup by properly scaling up the number of nodes $n$, where the speedup will asymptotically converge to $\frac{h}{e}$.

\begin{table*}[!ht]
\centering{\small
\begin{tabular}{c c c c c}
\rowcolor{Gray} \toprule
 hidden size & number of layers & number of nodes & batch size & speedup \bigstrut\\
\midrule
\rowcolor{LightCyan} 6144 & 40 & 1 & 1024 & 1.91$\times$ \bigstrut\\
\midrule
8192 & 48 & 2 & 1536 & 1.75$\times$ \bigstrut\\
\midrule
\rowcolor{LightCyan} 10240 & 60 & 4 & 1792 & 1.63$\times$ \bigstrut\\
\midrule
12288 & 80 & 8 & 2304 & 1.55$\times$ \bigstrut\\
\midrule
\rowcolor{LightCyan} 16384 & 96 & 16 & 2176 & 1.46$\times$ \bigstrut\\
\midrule
20480 & 105 & 35 & 2528 & 1.46$\times$ \bigstrut\\
\midrule
\rowcolor{LightCyan} 25600 & 128 & 64 & 3072 & 1.47$\times$\bigstrut\\
\bottomrule
\end{tabular}}
\vspace{-1mm}
\caption{
Weak-scaling speedup for the Transformer models. The number of tensor model parallelism is 4, and the micro-batch size is 16. As for the change of the hidden size, the number of layers, and the batch size, we follow the setting of Table 1 in~\cite{narayanan2021efficient}.
}
\vspace{-4mm}
\label{tab:up_cluster}
\end{table*}

In summary, compression in model parallelism has diminishing returns if we only scale up the model on a fixed cluster. To gain benefits from compression methods, one needs to also \textbf{properly manage other parameters in the cost model, e.g. also scaling up the number of nodes and use the pipeline parallelism}.
\section{Related Work}
\label{sec:related}
In this section, we first introduce work related to the development of large Transformer models. Then, we discuss strategies to train these models at scale. In the end, we discuss prior work that accelerates distributed ML models training by using compression techniques.

\noindent \textbf{Transformer Models.} Transformer models were first introduced by~\citet{vaswani2017attention} in the machine translation context. It has been shown to be effective in various other language understanding tasks such as text generation, text classification and question answering~\citep{devlin2018bert, radford2018improving, wang2018glue, rajpurkar2016squad}. Recent research has also successfully applied Transformer models to images~\citep{dosovitskiy2020image, touvron2021training}, audio~\citep{gong2021ast} and beyond~\citep{sharir2021image}. An $N$-layers  transformer model is composed of three major components: (1) An embedding layer that maps an input token to a hidden state, (2) A stack of $N$ transformer layers, and (3) a prediction layer that maps the hidden state proceeded by transformer layers to the task output. A transformer layer is composed of an attention module~\cite{bahdanau2014neural} and several matrix multiplications. Several optimizations have been proposed to speed up Transformer model training such as carefully managing the I/O~\cite{dao2022flashattention} and reducing the complexity of the attention module~\cite{wang2020linformer}. In this work, we speed up the Transformer model training in the \textit{distributed} setting, where we reduce the communication between workers.

\noindent\textbf{Training Large Transformer models.}
Several parallelism strategies have been proposed to train Transformer models. Megatron~\citep{shoeybi2019megatron} proposes tensor model parallelism, which parallelizes the computation in attention layers and in the following matrix multiplications. DeepSpeed~\citep{rasley2020deepspeed} uses a specialized form of pipeline parallelism~\citep{huang2019gpipe, narayanan2019pipedream} that treats a transformer layer as the smallest unit in pipeline stages. It further combines the tensor model parallelism developed in Megatron and data parallelism to train Transformer models at the scale of trillion parameters.~\citep{li2022amp} considers a more sophisticated model parallelism strategy space for Transformer models and uses a cost model to automatically search for the optimal one. Our work is orthogonal to the direction of developing new parallel training strategies. In this work, we study how to compress communication on existing parallel strategies.

\noindent \textbf{Distributed training with Compression.}
Distributed ML model training requires frequent and heavy synchronization between workers. Several directions have been proposed to reduce the communication bottleneck by compressing the message size. One direction is developed on the data parallelism setting, where workers communicate model gradients~\cite{wang2021pufferfish, agarwal2022utility} during backward propagation. Common techniques to reduce the gradient communication include low-rank updates~\cite{wang2018atomo}, sparsification~\cite{lin2017deep}, and quantization~\cite{seide20141,bernstein2018signsgd,dettmers20158}. A more recent direction find that the 
activation produced during the forward propagation in neural networks is large, and thus compressing them is beneficial~\cite{wang2022fine}. In particular, they use quantization to compress the activation volume between pipeline parallelism workers. However, they focus on the geo-distributed setting where the network bandwidth is very low. In this paper, we study the effect of a rich set of popular compression techniques on tensor and pipeline parallelism, and in a typical cloud computing setting.

\section{Conclusion}
In this work, we studied the impact of compressing activations for models trained using model parallelism. We implemented and integrated several popular compression algorithms into an existing distributed training framework (Megatron-LM) and evaluated their performance in terms of throughput and accuracy under various settings. Our results show that learning-based compression algorithms are the most effective approach for compressing activations in model parallelism. We also developed a performance model to analyze the speedup when scaling up the model. Our experiments provide valuable insights for the development of improved activation compression algorithms in the future.

\subsection*{Acknowledgments}
Shivaram Venkataraman is supported by the Office of the Vice Chancellor for Research and Graduate Education at UW-Madison with funding from the Wisconsin Alumni Research Foundation. Eric Xing is supported by NSF IIS1563887, NSF CCF1629559, NSF IIS1617583, NGA HM04762010002, NSF IIS1955532, NSF CNS2008248, NSF IIS2123952, and NSF BCS2040381.
\nocite{langley00}

\bibliography{example_paper}
\bibliographystyle{mlsys2023}

\newpage
\appendix
\onecolumn
\section{More Experimental Results}
\label{sec:more_exp_res}
We provide more experimental results in this section.

\begin{table*}[!ht]
\centering
\begin{tabular}{c c c c c c c c}
\toprule
Distributed Setting & w/o & A1 & A2 & T1 & T2 & T3 & T4 \\ 
\hline
TP=1, PP=4 & 151.82 & 154.62 & 155.03 & 155.78 & 155.12 & 156.84 & 158.58 \\ 
TP=2, PP=2 & 145.58 & 157.49 & 163.63 & 175.67 & 177.39 & 186.71 & 178.91 \\ 
TP=4, PP=1 & 136.66 & 155.43 & 145.97 & 170.04 & 176.88 & 186.06 & 190.01 \\ 
\bottomrule
Distributed Setting & R1 & R2 & R3 & R4 & Q1 & Q2 & Q3 \\
\hline
TP=1, PP=4 & 206.89 & 273.49 & 449.70 & 1,292.15 & 154.30 & 153.65 & 152.33 \\ 
TP=2, PP=2 & 844.66 & 1,589.66 & 3,915.32 & 15,732.57 & 178.09 & 175.23 & 172.93 \\ 
TP=4, PP=1 & 820.37 & 1,588.59 & 3,915.52 & 15,469.87 & 188.10 & 168.90 & 167.90 \\ 
\bottomrule
\end{tabular}
\caption{The total time (ms) for fine-tuning with various compression techniques by varying the distributed setting. The results are collected from the AWS p3.8xlarge machine with NVLink by using batch size 32, and sequence length 128.}
\label{tab:run_time_finetune_AWS_32_128}
\end{table*}


\begin{table*}[!ht]
\centering
\begin{tabular}{c c c c c c c c}
\toprule
Distributed Setting & w/o & A1 & A2 & T1 & T2 & T3 & T4 \\ 
\hline
TP=1, PP=4 & 106.04 & 113.67 & 106.35 & 109.58 & 109.10 & 109.18 & 110.57 \\ 
TP=2, PP=2 & 121.26 & 142.41 & 140.05 & 152.91 & 154.60 & 162.00 & 157.12 \\ 
TP=4, PP=1 & 122.22 & 142.33 & 139.47 & 171.24 & 165.77 & 172.69 & 170.61 \\ 
\bottomrule
Distributed Setting & R1 & R2 & R3 & R4 & Q1 & Q2 & Q3 \\
\hline
TP=1, PP=4 & 124.39 & 137.51 & 187.59 & 333.61 & 108.18 & 109.56 & 109.49 \\ 
TP=2, PP=2 & 314.51 & 507.00 & 998.51 & 3,197.42 & 163.18 & 155.48 & 150.31 \\ 
TP=4, PP=1 & 329.33 & 513.89 & 1,007.65 & 3,406.20 & 171.06 & 163.96 & 152.82 \\ 
\bottomrule
\end{tabular}
\caption{The total time (ms) for fine-tuning with various compression techniques by varying the distributed setting. The results are collected from the AWS p3.8xlarge machine with NVLink by using batch size 8, and sequence length 128.}
\label{tab:run_time_finetune_AWS_8_128}
\end{table*}

\begin{table*}[!ht]
\centering
\begin{tabular}{c c c c c c c c}
\toprule
Distributed Setting & w/o & A1 & A2 & T1 & T2 & T3 & T4 \\ 
\hline
TP=1, PP=4 & 154.82 & 152.50 & 153.47 & 155.56 & 156.01 & 156.81 & 158.37 \\ 
TP=2, PP=2 & 184.48 & 175.29 & 180.35 & 206.56 & 204.48 & 207.66 & 214.30 \\ 
TP=4, PP=1 & 212.76 & 201.39 & 200.31 & 234.16 & 240.42 & 242.62 & 261.39 \\ 
\bottomrule
Distributed Setting & R1 & R2 & R3 & R4 & Q1 & Q2 & Q3 \\
\hline
TP=1, PP=4 & 185.83 & 231.78 & 368.95 & 963.62 & 155.33 & 154.85 & 154.82 \\ 
TP=2, PP=2 & 684.28 & 1,228.36 & 2,900.86 & 10,499.14 & 188.82 & 189.14 & 194.25 \\ 
TP=4, PP=1 & 722.87 & 1,275.57 & 2,973.04 & 10,891.70 & 225.42 & 230.69 & 242.42 \\ 
\bottomrule
\end{tabular}
\caption{The total time (ms) for fine-tuning with various compression techniques by varying the distributed setting. The results are collected from the local machine without NVLink by using batch size 32, and sequence length 128.}
\label{tab:run_time_finetune_local_32_128}
\end{table*}


\begin{table*}[!ht]
\centering
\begin{tabular}{c c c c c c c c}
\toprule
Distributed Setting & w/o & A1 & A2 & T1 & T2 & T3 & T4 \\ 
\hline
TP=1, PP=4 & 73.19 & 72.94 & 72.58 & 75.98 & 74.15 & 73.62 & 74.86 \\ 
TP=2, PP=2 & 100.86 & 107.73 & 100.54 & 113.59 & 117.36 & 114.86 & 112.11 \\ 
TP=4, PP=1 & 100.73 & 107.90 & 115.18 & 129.31 & 124.94 & 136.18 & 133.91 \\ 
\bottomrule
Distributed Setting & R1 & R2 & R3 & R4 & Q1 & Q2 & Q3 \\
\hline
TP=1, PP=4 & 82.45 & 94.84 & 123.78 & 239.81 & 73.33 & 74.41 & 71.80 \\ 
TP=2, PP=2 & 235.02 & 366.59 & 769.47 & 2,183.39 & 111.61 & 106.75 & 101.25 \\ 
TP=4, PP=1 & 238.28 & 368.45 & 733.03 & 2,509.73 & 120.14 & 114.73 & 118.98 \\ 
\bottomrule
\end{tabular}
\caption{The total time (ms) for fine-tuning with various compression techniques by varying the distributed setting. The results are collected from the local machine without NVLink by using batch size 8, and sequence length 128.}
\label{tab:run_time_finetune_local_8_128}
\end{table*}

\begin{table*}[!ht]
\centering
\begin{tabular}{c c c c c c c c c}
\toprule
\makecell[c]{Compression \\ Algorithm} & MNLI-(m/mm) & QQP & SST-2 & MRPC & CoLA & QNLI & RTE & STS-B \\ 
\hline
w/o & 87.87/88.02 & 91.96 & 95.18 & 87.71 & 59.40 & 92.99 & 76.90 &  88.43 \\ 
\hline
A1 & 85.30/85.33 & 91.28 & 92.32 & 84.58 & 55.18 & 90.87 & 59.93 & 87.92  \\ 
A2 & 85.25/85.19 & 91.41 & 93.23 & 86.72 & 57.02 & 90.92 & 64.26 & 87.74  \\ 
\hline
T1 & 34.38/34.01 & 72.29 & 49.54 & 70.38 & 36.64 & 59.89 & 53.43 & 70.81  \\ 
T2 & 40.10/38.97 & 58.91 & 79.24 & 66.49 & 0.00 & 80.40 & 45.49 & 11.32  \\ 
T3 & 68.76/69.23 & 64.58 & 91.40 & 80.93 & 0.00 & 67.34 & 66.43 & 69.24  \\
T4 & 84.24/85.23 & 89.17 & 92.09 & 81.68 & 51.54 & 91.71 & 63.54 & 84.80  \\ 
\hline
Q1 & 86.85/87.58 & 91.50 & 93.58 & 86.96 & 59.20 & 92.24 & 59.57 & 86.89  \\ 
Q2 & 87.46/88.02 & 91.82 & 94.95 & 87.48 & 57.02 & 93.36 & 68.95 & 87.84  \\ 
\bottomrule
\end{tabular}
\caption{Fintune results over GLUE dataset under the setting using tensor parallelism size 2, pipeline parallelism size 2, batch size 32, and sequence length 128. F1 scores are reported for QQP and MRPC, Matthews correlation coefficient is reported for CoLA, and Spearman correlations are reported for STS-B, and accuracy scores are reported for the other tasks.}
\label{tab:accuracy_finetune_32_128}
\end{table*}


\begin{table*}[!ht]
\centering
\begin{tabular}{c c c c c c c c c}
\toprule
\makecell[c]{Compression \\ Algorithm} & MNLI-(m/mm) & QQP & SST-2 & MRPC & CoLA & QNLI & RTE & STS-B \\ 
\hline
w/o & 86.23/86.07 & 91.22 & 91.74 & 88.17 & 59.02 & 92.09 & 78.70 & 88.40  \\ 
\hline
A1 & 82.49/82.41 & 89.93 & 91.85 & 82.43 & 43.56 & 89.84 & 47.29 & 87.03  \\ 
A2 & 82.18/82.23 & 90.45 & 90.52 & 83.54 & 0.00 & 89.02 & 62.82 & 87.66  \\ 
\hline
T1 & 36.69/38.13 & 66.85 & 55.32 & 68.93 & 0.00 & 59.13 & 52.71 & 1.97  \\ 
T2 & 43.92/43.66 & 73.63 & 51.26 & 62.26 & 0.00 & 60.13 & 49.82 & 0.00  \\ 
T3 & 49.07/47.96 & 72.02 & 83.57 & 69.33 & 12.04 & 83.60 & 55.60 & 84.96  \\ 
T4 & 83.99/84.37 & 35.78 & 68.30 & 83.54 & 47.33 & 60.52 & 64.62 & 86.72  \\ 
\hline
Q1 & 84.91/85.18 & 90.54 & 92.43 & 85.91 & 53.25 & 60.68 & 57.04  & 87.91  \\ 
Q2 & 85.66/86.09 & 90.99 & 91.74 & 86.84 & 53.92 & 91.31 & 75.81 & 88.19  \\ 
\bottomrule
\end{tabular}
\caption{Fintune results over GLUE dataset under the setting using tensor parallelism size 2, pipeline parallelism size 2, batch size 8, and sequence length 128. F1 scores are reported for QQP and MRPC, Matthews correlation coefficient is reported for CoLA, and Spearman correlations are reported for STS-B, and accuracy scores are reported for the other tasks.}
\label{tab:accuracy_finetune_8_128}
\end{table*}

%


\end{document}